\documentclass[10pt,twocolumn,letterpaper]{article}

\usepackage{cvpr}              

\usepackage{adjustbox}
\usepackage{multirow}
\usepackage{tabularx}

\usepackage{amssymb}
\usepackage{makecell}
\usepackage[accsupp]{axessibility} 
\usepackage{tikz}
\usetikzlibrary{positioning,arrows.meta,fit,backgrounds}

\usepackage{longtable}

\definecolor{cvprblue}{rgb}{0.21,0.49,0.74}
\usepackage[pagebackref,breaklinks,colorlinks,citecolor=cvprblue]{hyperref}

\def\paperID{NTIRE-142} 
\def\confName{CVPR}
\def\confYear{2026}

\title{The Second Challenge on Real-World Face Restoration at NTIRE 2026: Methods and Results}

\author{Jingkai Wang\thanks{\href{mailto:jingkaiwang100@gmail.com}{Jingkai Wang}, \href{mailto:g1017325431@gmail.com}{Jue Gong}, \href{mailto:zhengchen.cse@gmail.com}{Zheng Chen}, \href{mailto:normal.kliu@gmail.com}{Kai Liu}, \href{mailto:jiatong.li2024@gmail.com}{Jiatong Li},
\href{mailto:yulun100@gmail.com}{Yulun Zhang}, and \href{mailto:Radu.Timofte@uni-wuerzburg.de}{Radu Timofte} are the challenge organizers, while the other authors participated in the challenge. Section~B in the supplementary materials contains the authors' teams and affiliations. NTIRE 2026 webpage: \url{https://cvlai.net/ntire/2026}. Code: \url{https://github.com/jkwang28/NTIRE2026_RealWorld_Face_Restoration}. } \and Jue Gong\footnotemark[1] \and Zheng Chen\footnotemark[1] \and Kai Liu\footnotemark[1] \and Jiatong Li\footnotemark[1] \and Yulun Zhang\footnotemark[1] \thanks{Corresponding author: Yulun Zhang. \href{mailto:yulun100@gmail.com}{yulun100@gmail.com}} \and Radu Timofte\footnotemark[1] \and
Jiachen Tu \and Yaokun Shi \and Guoyi Xu \and Yaoxin Jiang \and Jiajia Liu \and 
Yingsi Chen \and Yijiao Liu \and Hui Li \and Yu Wang \and Congchao Zhu \and
Alexandru-Gabriel Lefterache \and Anamaria Radoi \and
Chuanyue Yan \and Tao Lu \and Yanduo Zhang \and Kanghui Zhao \and Jiaming Wang \and Yuqi Li \and
WenBo Xiong \and Yifei Chen \and Xian Hu \and Wei Deng \and Daiguo Zhou \and
Sujith Roy V \and Claudia Jesuraj \and Vikas B \and Spoorthi LC \and Nikhil Akalwadi \and Ramesh Ashok Tabib \and Uma Mudenagudi \and
Yuxuan Jiang \and Chengxi Zeng \and Tianhao Peng \and Fan Zhang \and David Bull
Wei Zhou \and Linfeng Li \and Hongyu Huang \and
Hoyoung Lee \and SangYun Oh \and ChangYoung Jeong \and
Axi Niu \and Jinyang Zhang \and Zhenguo Wu \and Senyan Qing \and Jinqiu Sun \and Yanning Zhang
}

\begin{document}

\maketitle

\begin{abstract}
This paper provides a review of the NTIRE 2026 challenge on real-world face restoration, highlighting the proposed solutions and the resulting outcomes. The challenge focuses on generating natural and realistic outputs while maintaining identity consistency. Its goal is to advance state-of-the-art solutions for perceptual quality and realism, without imposing constraints on computational resources or training data. Performance is evaluated using a weighted image quality assessment (IQA) score and employs the AdaFace model as an identity checker. The competition attracted 96 registrants, with 10 teams submitting valid models; ultimately, 9 teams achieved valid scores in the final ranking. This collaborative effort advances the performance of real-world face restoration while offering an in-depth overview of the latest trends in the field.
\end{abstract}

\vspace{-4mm}
\section{Introduction}
\vspace{-1mm}
Face restoration aims to reconstruct high-quality (HQ) face images from low-quality (LQ) inputs degraded by blur, noise, compression, and other distortions. Since severe degradation often removes a large amount of visual information, this task is inherently ill-posed. Meanwhile, with the continuous progress of portrait imaging technology, users increasingly expect restored face images to exhibit both rich details and high fidelity. This makes it essential for restoration methods to produce outputs that are not only clear but also natural and realistic. In recent years, deep learning has substantially advanced face restoration. Methods based on CNNs, Transformers~\cite{zhou2022codeformer,wang2023restoreformer++,xie2024pltrans,tsai2024daefr}, GANs~\cite{ChenPSFRGAN,wang2021gfpgan,Yang2021GPEN,chan2021glean}, and diffusion models~\cite{wang2023dr2,miao2024waveface,yang2023pgdiff,chen2023BFRffusion,qiu2023diffbfr,Suin2024CLRFace,wu2024osediff,diffbir,yue2024difface,tao2025overcoming,wang2025osdface} demonstrated strong performance.

A key challenge in this field lies in how to effectively model face priors. Traditional image restoration methods often rely on statistical priors, whereas modern neural methods tend to learn such priors directly from data. Among them, geometric-prior-based approaches~\cite{yu2018super, chen2018fsrnet, kim2019progressive, shen2018deep} are particularly valuable because they provide explicit structural cues for facial reconstruction. However, when the degradation is relatively mild, users often expect the restored results to remain highly realistic, including subtle skin textures that are usually captured only by high-end imaging devices. Therefore, beyond semantic guidance, texture priors are also vital for recovering fine facial details.

Recent studies~\cite{gu2022vqfr,zhou2022codeformer,wang2023restoreformer++,xie2024pltrans,tsai2024daefr} have extensively explored Transformer-based designs for incorporating face priors. Representative methods such as CodeFormer~\cite{zhou2022codeformer} and DAEFR~\cite{tsai2024daefr} employ codebooks learned from HQ face images as priors. Although these methods are effective at preserving facial information, they still show limitations when handling severely degraded images, especially in transition regions between the face and the background.

For more severely degraded inputs, generative capability becomes increasingly important. GAN-based methods~\cite{ChenPSFRGAN,wang2021gfpgan,Yang2021GPEN,chan2021glean} have shown strong ability in synthesizing plausible facial details. Among them, GFPGAN~\cite{wang2021gfpgan} is particularly notable, not only for its effective restoration framework, but also for providing benchmark datasets widely used by the computer vision community. More recently, diffusion-based methods~\cite{wang2023dr2,miao2024waveface,yang2023pgdiff,chen2023BFRffusion,qiu2023diffbfr,Suin2024CLRFace,wu2024osediff,diffbir,yue2024difface,tao2025overcoming,wang2025osdface} have emerged as a powerful paradigm. Benefiting from the strong generative priors of diffusion models, high-quality face restoration from severely degraded inputs has become increasingly feasible. DR2~\cite{wang2023dr2} transforms the input into noisy states and progressively denoises it to recover essential semantic information. DiffBIR~\cite{diffbir} further improves facial detail restoration by leveraging a pre-trained latent diffusion model as a strong prior. In addition, super-resolution models such as SUPIR~\cite{yu2024scaling} and StableSR~\cite{wang2024exploiting} have also been widely adopted in this competition, further highlighting the effectiveness of diffusion-based techniques for real-world face restoration.

Very recently, researchers have made significant progress in advancing the field of face restoration. 
FaceMe~\cite{liu_faceme_2025} and RefSTAR~\cite{yin_refstar_2026} combine reference images with diffusion models, greatly improving reference-based face restoration. 
InterLCM~\cite{li_interlcm_2025} introduces latent consistency models to the field. It uses a 4-step LCM to improve inference efficiency.
OSDFace~\cite{wang2025osdface} uses pre-trained models to reduce multi-step diffusion sampling to a single step. This achieves faster inference while maintaining high restoration quality. 
\citep{qiu_feature_2025} uses the Schrödinger Bridge and Pseudo-Hashing to explore optimal transport paths during face restoration.
FLIPNET~\cite{miao_unlocking_nodate} integrates restoration and degradation modes, which offer a new paradigm for learning real-world degradation.
SSDiff~\cite{li_self-supervised_2025} focuses on old photo restoration. It proposes a training-free method that uses a staged, region-specific guidance scheme.

In collaboration with the 2026 New Trends in Image Restoration and Enhancement (NTIRE 2026) workshop, we organized a challenge on real-world face restoration. The challenge aims to recover high-quality face images from degraded low-quality inputs, with an emphasis on richer textures, more realistic facial appearances, and consistent identity preservation. Its goal is to encourage the development of solutions that achieve strong restoration quality with the best perceptual performance, while also revealing current trends in face restoration design.


This challenge is one of the challenges associated with the NTIRE 2026 Workshop~\footnote{\url{https://www.cvlai.net/ntire/2026/}} on:
deepfake detection~\cite{ntire26deepfake}, 
high-resolution depth~\cite{ntire26hrdepth},
multi-exposure image fusion~\cite{ntire26raim_fusion}, 
AI flash portrait~\cite{ntire26raim_portrait}, 
professional image quality assessment~\cite{ntire26raim_piqa},
light field super-resolution~\cite{ntire26lightsr},
3D content super-resolution~\cite{ntire263dsr},
bitstream-corrupted video restoration~\cite{ntire26videores},
X-AIGC quality assessment~\cite{ntire26XAIGCqa},
shadow removal~\cite{ntire26shadow},
ambient lighting normalization~\cite{ntire26lightnorm},
controllable Bokeh rendering~\cite{ntire26bokeh},
rip current detection and segmentation~\cite{ntire26ripdetseg},
low light image enhancement~\cite{ntire26llie},
high FPS video frame interpolation~\cite{ntire26highfps},
Night-time dehazing~\cite{ntire26nthaze,ntire26nthaze_rep},
learned ISP with unpaired data~\cite{ntire26isp},
short-form UGC video restoration~\cite{ntire26ugcvideo},
raindrop removal for dual-focused images~\cite{ntire26dual_focus},
image super-resolution (x4)~\cite{ntire26srx4},
photography retouching transfer~\cite{ntire26retouching},
mobile real-word super-resolution~\cite{ntire26rwsr},
remote sensing infrared super-resolution~\cite{ntire26rsirsr},
AI-Generated image detection~\cite{ntire26aigendet},
cross-domain few-shot object detection~\cite{ntire26cdfsod},
financial receipt restoration and reasoning~\cite{ntire26finrec},
real-world face restoration~\cite{ntire26faceres},
reflection removal~\cite{ntire26reflection},
anomaly detection of face enhancement~\cite{ntire26anomalydet},
video saliency prediction~\cite{ntire26videosal},
efficient super-resolution~\cite{ntire26effsr},
3d restoration and reconstruction in adverse conditions~\cite{ntire26realx3d},
image denoising~\cite{ntire26denoising},
blind computational aberration correction~\cite{ntire26aberration},
event-based image deblurring~\cite{ntire26eventblurr},
efficient burst HDR and restoration~\cite{ntire26bursthdr},
low-light enhancement: `twilight cowboy'~\cite{ntire26twilight},
and efficient low light image enhancement~\cite{ntire26effllie}.

\vspace{-1mm}
\section{NTIRE 2026 Real-world Face Restoration}
\vspace{-0.5mm}
This challenge focuses on restoring real-world degraded face images. The task is to recover high-quality face images with rich high-frequency details from low-quality inputs. At the same time, the output should preserve facial identity to a reasonable degree. There are no restrictions on computational resources such as model size or FLOPs. The main goal is to achieve the best possible image quality and identity consistency.

\begin{table*}[t]
\centering
\small
\begin{adjustbox}{width=\linewidth}
\begin{tabular}{cc|c|cccccc|ccc|c}
\toprule
\multirow{2}{*}{\shortstack{Team\\No.}} & \multirow{2}{*}{\shortstack{Team\\Name}} & \multirow{2}{*}{Rank} & \multirow{2}{*}{NIQE} & \multirow{2}{*}{CLIPIQA} & \multirow{2}{*}{ManIQA} & \multirow{2}{*}{MUSIQ} & \multirow{2}{*}{Q-Align} & \multirow{2}{*}{FID} & \multirow{2}{*}{\shortstack{Adaface\\Score}} & \multirow{2}{*}{\shortstack{Failed\\images}} & \multirow{2}{*}{\shortstack{ID\\Validation}} & \multirow{2}{*}{\shortstack{Total\\Score}} \\
& & & & & & & & & & & & \\
\midrule
5  & MiPlusCV       & 1   & 3.6897 & 0.9346 & 0.9082 & 77.5060 & 4.4648 & 53.6291 & 0.8273 & 1  & \checkmark & 4.6055 \\
6  & KLETech-CEVI   & 2   & 3.5486 & 0.9537 & 0.6563 & 77.3132 & 4.3771 & 56.0744 & 0.8425 & 1  & \checkmark & 4.3429 \\
2  & HONORAICamera  & 3   & 3.8173 & 0.9343 & 0.6026 & 76.1874 & 4.3017 & 52.6388 & 0.7770 & 0  & \checkmark & 4.2510 \\
8  & YuFans         & 4   & 3.8743 & 0.9655 & 0.6519 & 78.0809 & 3.9467 & 56.1398 & 0.8524 & 1  & \checkmark & 4.2387 \\
10 & guaguagua      & 5   & 3.9447 & 0.7327 & 0.6027 & 76.1343 & 4.4872 & 52.2211 & 0.7636 & 5  & \checkmark & 4.0775 \\
1  & NTR            & 6   & 4.9374 & 0.7638 & 0.6064 & 75.2589 & 4.4409 & 61.6428 & 0.6801 & 10 & \checkmark & 3.9008 \\
3  & MaDENN         & 7   & 4.3231 & 0.7020 & 0.5293 & 74.9940 & 4.2022 & 53.1356 & 0.7834 & 0  & \checkmark & 3.8581 \\
9  & SN\_VISION     & 8   & 7.1777 & 0.7563 & 0.5286 & 67.5346 & 3.4691 & 67.5720 & 0.7325 & 5  & \checkmark & 3.2606 \\
4  & ALLCAN         & 9   & 6.1302 & 0.5672 & 0.4583 & 62.1487 & 3.6028 & 74.7110 & 0.7509 & 2  & \checkmark & 3.0075 \\
\midrule
7  & BVI            & N/A & 4.3872 & 0.8193 & 0.6857 & 77.0376 & 4.6288 & 66.7819 & 0.5334 & 82 & $\times$ & 4.0946 \\
\midrule
\end{tabular}
\end{adjustbox}
\caption{Results of NTIRE 2026 Real-world Face Restoration Challenge. The testing was conducted on the test dataset, consisting of 450 images from CelebChild-Test, LFW-Test, WIDER-Test, CelebA, and WebPhoto-Test. Participants were required to pass the AdaFace ID Test first to qualify for ranking. The final results were calculated based on the weighted score of no-reference IQA metrics for the ranking.}
\label{tab:main_results}
\end{table*}

\subsection{Dataset}
We recommend FFHQ~\cite{karras2019ffhq} as the primary training dataset, which provides 70,000 HQ face images. Participants may also use other datasets during training. Separate image sets are adopted for the development and evaluation phases. Specifically, the test set consists of images sampled from five datasets, including 50 images from CelebChild-Test~\cite{wang2021gfpgan}, and 100 images each from LFW-Test~\cite{wang2021gfpgan}, WIDER-Test~\cite{zhou2022codeformer}, CelebA~\cite{karras2018celeba}, and WebPhoto-Test~\cite{wang2021gfpgan}.

\vspace{0.5mm}
\noindent{\textbf{FFHQ.}} FFHQ consists of 70,000 high-quality face images covering diverse identities, attributes, and demographic characteristics. Owing to its high resolution and strong consistency, it is commonly adopted for face generation and restoration tasks.

\vspace{0.5mm}
\noindent{\textbf{LFW-Test.}} LFW-Test is constructed from the Labeled Faces in the Wild (LFW) dataset~\cite{huang2008lfw} and contains 1,711 low-quality face images captured in unconstrained environments. It is built by taking the first image of each identity from the validation split.

\vspace{0.5mm}
\noindent{\textbf{WIDER-Test.}} WIDER-Test includes 970 low-quality real-world images sampled from the WIDER FACE dataset. The set covers challenging scenarios such as large pose variation, occlusion, and difficult lighting conditions.

\vspace{0.5mm}
\noindent{\textbf{CelebChild-Test.}} CelebChild-Test contains 180 childhood celebrity face images collected from online sources. Many samples are black-and-white or of limited quality, representing severe real-world degradation.

\vspace{0.5mm}
\noindent{\textbf{WebPhoto-Test.}} WebPhoto-Test is derived from 188 real-world images collected from the Internet, from which 407 faces are detected. It presents complex degradations, including aging effects, detail degradation, and color fading.

\vspace{0.5mm}
\noindent{\textbf{CelebA.}} In this challenge, CelebA is sampled from the validation split of the CelebFaces Attributes (CelebA) dataset~\cite{karras2018celeba}, which contains 19,867 images with a resolution of 178$\times$218. All images are first center-cropped to 178$\times$178 and then resized to 512$\times$512.

\subsection{Competition} \label{sec:evaluation}
Participants are ranked according to the visual quality of their restored face images, while preserving identity consistency with the corresponding low-quality input faces in the test set. Submissions must keep identity similarity above a predefined threshold, with at most 10 cases falling below the threshold, and should further pursue the highest possible perceptual quality scores.

\subsubsection{Challenge Phases}
\noindent\textbf{Development and Validation Phase:} Participants are given 70,000 high-quality images from the FFHQ dataset, together with 450 low-quality (LQ) images sampled from five real-world datasets. By introducing simulated degradations, they can build paired training data for supervised face restoration. The use of additional training datasets is also allowed. During this Phase, participants may submit their restored high-quality images to the CodaBench evaluation server and obtain perceptual quality scores, including CLIPIQA~\cite{wang2022clipiqa} and MUSIQ~\cite{ke2021musiq}.

\vspace{1mm}
\noindent\textbf{Testing Phase:} In the final testing phase, participants receive another set of 450 low-quality test images that are different from those used in development, which is also different from the former challenges. They are also required to upload their restored results to the CodaBench server and send their code together with a factsheet to the organizers by email. The organizers will then validate the submitted code and release the final rankings after the challenge is finished.

\subsubsection{Evaluation Procedure}
\noindent\textbf{Step 1: Identity Similarity Measurement.} We adopt a pre-trained AdaFace~\cite{kim2022adaface} model to extract identity embeddings from the input low-quality images and the restored high-quality (HQ) images, and then measure their cosine similarity. Since the severity of degradation varies across datasets, different identity thresholds are used for different data sources. The threshold is set to 0.3 for Wider-Test and WebPhoto-Test, 0.6 for LFW-Test and CelebChild-Test, and 0.5 for CelebA.

\vspace{0.5mm}

\noindent\textbf{Step 2: Image Quality Assessment.} The restored HQ images are assessed with several no-reference image quality assessment (IQA) metrics, including NIQE~\cite{zhang2015niqe}, CLIPIQA~\cite{wang2022clipiqa}, MANIQA~\cite{yang2022maniqa}, MUSIQ~\cite{ke2021musiq}, and Q-Align~\cite{wu2024qalign}. We also compute the FID score using FFHQ as the reference dataset. To ensure fairness and reproducibility, the final ranking is primarily determined by the results reproduced from the submitted code, which are used for verification. The CodaBench submission is used as a supplementary reference, and small score discrepancies are considered acceptable. The evaluation scripts are publicly available at {\url{https://github.com/jkwang28/NTIRE2026_RealWorld_Face_Restoration}}, where the source code and pre-trained models of participating methods are also provided.
The teams are ultimately ranked based on the overall perceptual score, which is computed by
\vspace{-2mm}\begin{equation*}
\begin{aligned}
   \text{Score} &= \text{CLIPIQA} + \text{MANIQA} + \frac{\text{MUSIQ}}{100} + \frac{\text{QALIGN}}{5}\\
   & + \max\left(0, \frac{10 - \text{NIQE}}{10}\right) + \max\left(0, \frac{100-\text{FID}}{100}\right). 
\end{aligned}
\end{equation*}

\section{Challenge Results}

Table~\ref{tab:main_results} presents the final rankings and results of the teams. A comprehensive description of the evaluation process is outlined in Sec.~\ref{sec:evaluation}. All ten participating teams, together with their method details, are summarized in Sec.~\ref{sec:teams}. Team member information can be found in the appendix. MiPlusCV achieved first place in this year's challenge, followed by CEVI-KLETech, HONORAICamera, YuFans, and guaguagua. Only one team, BVI, failed the AdaFace ID test and therefore did not receive a valid final ranking.

\subsection{Architectures and main ideas}

Throughout this year's challenge, the strongest methods largely revolved around adapting powerful pre-trained image generators to the face restoration task. Based on the top-ranked teams in Table~\ref{tab:main_results}, we summarize the main ideas as follows.

\begin{enumerate}

    \item \textbf{One-step or distilled diffusion priors dominate the top ranks.}
    The first, third, and fourth-ranked teams all rely on strong one-step or fixed-timestep generative backbones. MiPlusCV combines OSDFace~\cite{wang2025osdface} with a Z-Image-based one-step diffusion restorer~\cite{zit}, HONORAICamera fine-tunes Z-Image-Turbo with OMGSR-style training~\cite{omgsr}, and YuFans directly builds on SDFace/OSDFace with an SDXL-Turbo prior~\cite{wang2025osdface,SDXL-Turbo}. This indicates that high perceptual quality can now be achieved with a single forward generation stage rather than only with expensive multi-step diffusion.

    \vspace{1mm}

    \item \textbf{Metric-oriented refinement becomes a key differentiator.}
    Several top teams do not stop at a strong base restorer, but explicitly optimize for the challenge metrics. MiPlusCV performs post-training refinement using CLIPIQA, MANIQA, and MUSIQ rewards, while YuFans applies direct CLIPIQA-guided pixel optimization at test time~\cite{wang2022clipiqa}. These results show that lightweight metric-aware refinement can produce clear leaderboard gains when the backbone already provides strong realism and identity preservation.

    \vspace{1mm}

    \item \textbf{Semantic and structural guidance remains essential for identity-safe restoration.}
    The second-ranked CEVI-KLETech method augments a three-stage baseline with semantic facial parsing and wavelet-domain correction, allowing different facial regions to receive different amounts of refinement. This is consistent with a broader trend in this year's submissions: even when the main generator is a large pre-trained diffusion model, explicit face-aware priors are still important for maintaining stable anatomy and identity.

    \vspace{1mm}

    \item \textbf{Modular multi-stage designs are still highly competitive.}
    Instead of training a single monolithic model end-to-end, the leading methods usually separate coarse recovery, perceptual enhancement, and optional post-processing. MiPlusCV uses a two-stage restoration pipeline, CEVI-KLETech inserts a lightweight correction module between diffusion and naturalness stages, and guaguagua adapts a large FLUX.2 model with degradation-aware structured control and LoRA. This modular design makes it easier to reuse strong generative priors while adding task-specific refinement blocks.

    \item \textbf{Foundation-model adaptation is replacing purely task-specific restoration backbones.}
    The top teams extensively build on large generative priors such as Z-Image, SDXL-Turbo, DiffBIR, and FLUX.2 rather than relying only on traditional face restoration networks. This year's challenge, therefore, highlights a clear shift toward adapting foundation image generators for perceptual face restoration, often with parameter-efficient tuning and task-specific control signals. 
\end{enumerate}

\vspace{-1.mm}
\subsection{Participants}
\vspace{-1.mm}
This year, the real-world face restoration challenge received 96 registrations, among which 10 teams submitted valid models. Following AdaFace-based identity verification, 9 teams remained eligible for the final ranking. Together, these submissions offer a representative view of current real-world face restoration methods operating under the dual requirements of perceptual quality and identity consistency.

\vspace{-1.mm}
\subsection{Fairness}
\vspace{-1mm}
To ensure a fair competition, we establish the following rules. \textbf{(1)} Participants are recommended to use the FFHQ dataset for training, and the training data must not contain any overlapping images with the five test datasets, namely LFW-Test, WIDER-Test, CelebChild-Test, CelebA, and WebPhoto-Test. \textbf{(2)} The use of additional training datasets, such as FFHQR, is allowed. \textbf{(3)} The use of no-reference IQA and simulated degradation pipelines in both training and testing is regarded as fair practice.

\vspace{-1.mm}
\subsection{Conclusions}
\vspace{-1.mm}

The main conclusions drawn from this year's challenge are summarized as follows:
\begin{enumerate}
    \item Perceptual face restoration is increasingly dominated by efficient generative paradigms, especially recent one-step and distilled approaches in practice.
    \item Strong results depend not only on the restoration backbone itself, but also on targeted refinement strategies such as semantic wavelet correction, metric-aware post-training, or test-time IQA optimization.
    \item Strong results do not rely on unconstrained generation alone, but combine foundation-model generation with semantic, structural, or identity-preserving constraints to preserve identity and facial structure.
\end{enumerate}

\begin{figure*}[t]
    \centering
    \includegraphics[width=\textwidth]{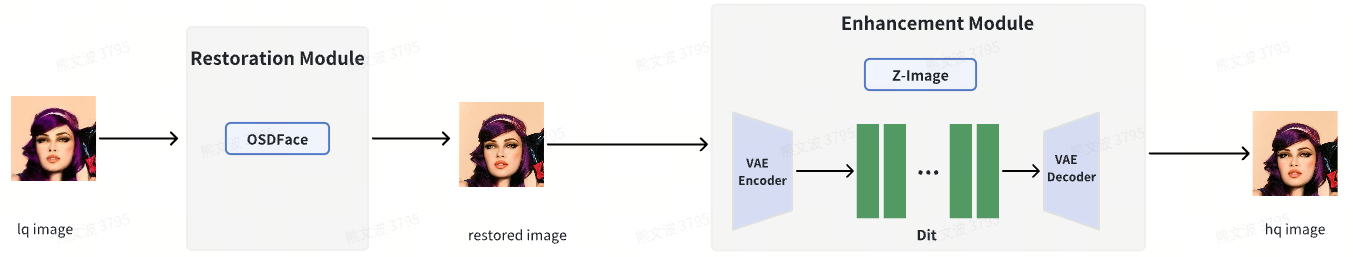}
    \caption{MiPlusCV adopts a two-stage pipeline that combines OSDFace-based coarse restoration with a Z-Image-based one-step detail enhancement stage.}
    \label{fig:team05-pipeline}
\end{figure*}

\section{Challenge Methods and Teams}
\label{sec:teams}
\subsection{MiPlusCV}
\noindent\textbf{Description.} MiPlusCV adopts a two-stage restoration framework. The first stage uses OSDFace~\cite{wang2025osdface} to recover coarse facial structure and suppress severe degradations, while the second stage refines facial details with a one-step diffusion restorer built on the pre-trained Z-Image foundation model~\cite{zit}. The design is tailored to achieve strong perceptual quality under the no-reference IQA metrics.

\vspace{1mm}
\noindent\textbf{Implementation Details.} The second stage is trained with LoRA adapters and direct image-level supervision rather than iterative diffusion sampling. Its objective combines an $\ell_1$ fidelity term, an edge-aware DISTS perceptual loss, ArcFace-based identity supervision~\cite{deng2019arcface}, and adversarial learning with a DINOv2-based discriminator~\cite{oquab2023dinov2}.

\vspace{1mm}
\noindent\textit{Training and optimization.} Shown in Fig.~\ref{fig:team05-pipeline}, the one-step model is optimized with AdamW using $\beta_1=0.5$, $\beta_2=0.999$, and a learning rate of $1\times10^{-4}$. Training uses FFHQ together with an additional 40,000 DSLR-captured high-resolution face images. After supervised training, the team further performs reward-based post-training using CLIPIQA, MANIQA, and MUSIQ as optimization signals.

\subsection{CEVI-KLETech}
\noindent\textbf{Description.} CEVI-KLETech proposes Semantic-Aware Frequency-Guided Residual Correction (SA-FGRC), a lightweight module inserted between a three-stage baseline composed of a StyleGAN2-based fidelity model~\cite{stylegan2}, DiffBIR~\cite{diffbir}, and a DINOv2-guided naturalness module~\cite{oquab2023dinov2}. Their core observation is that different facial regions require different amounts of high-frequency correction.

\begin{figure}[t]
    \centering
    \includegraphics[width=\linewidth]{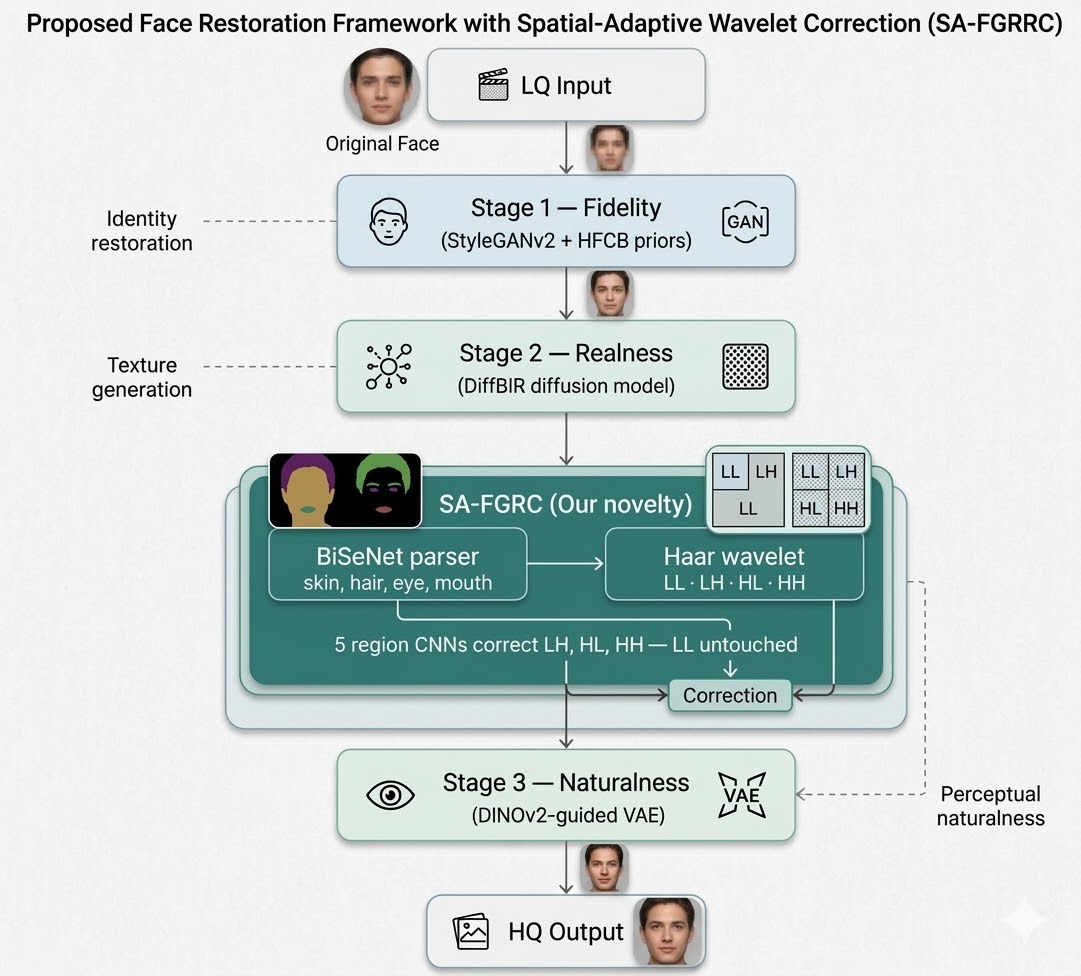}
    \caption{Overview of the CEVI-KLETech pipeline. A semantic-aware wavelet correction block is inserted between the diffusion and naturalness stages.}
    \label{fig:team06-pipeline} \vspace{-4mm}
\end{figure}

\vspace{1mm}
\noindent\textbf{Implementation Details.} SA-FGRC first decomposes the stage-2 restoration with a 2D Haar wavelet transform into one low-frequency band and three high-frequency bands. A BiSeNet parser~\cite{bisenet} then groups the face into skin, eyes, mouth, hair, and background, and five lightweight CNNs predict region-specific residual corrections for the high-frequency bands only. The low-frequency band remains untouched to preserve coarse structure and identity.

\begin{figure}[t]
    \centering
    \includegraphics[width=\linewidth]{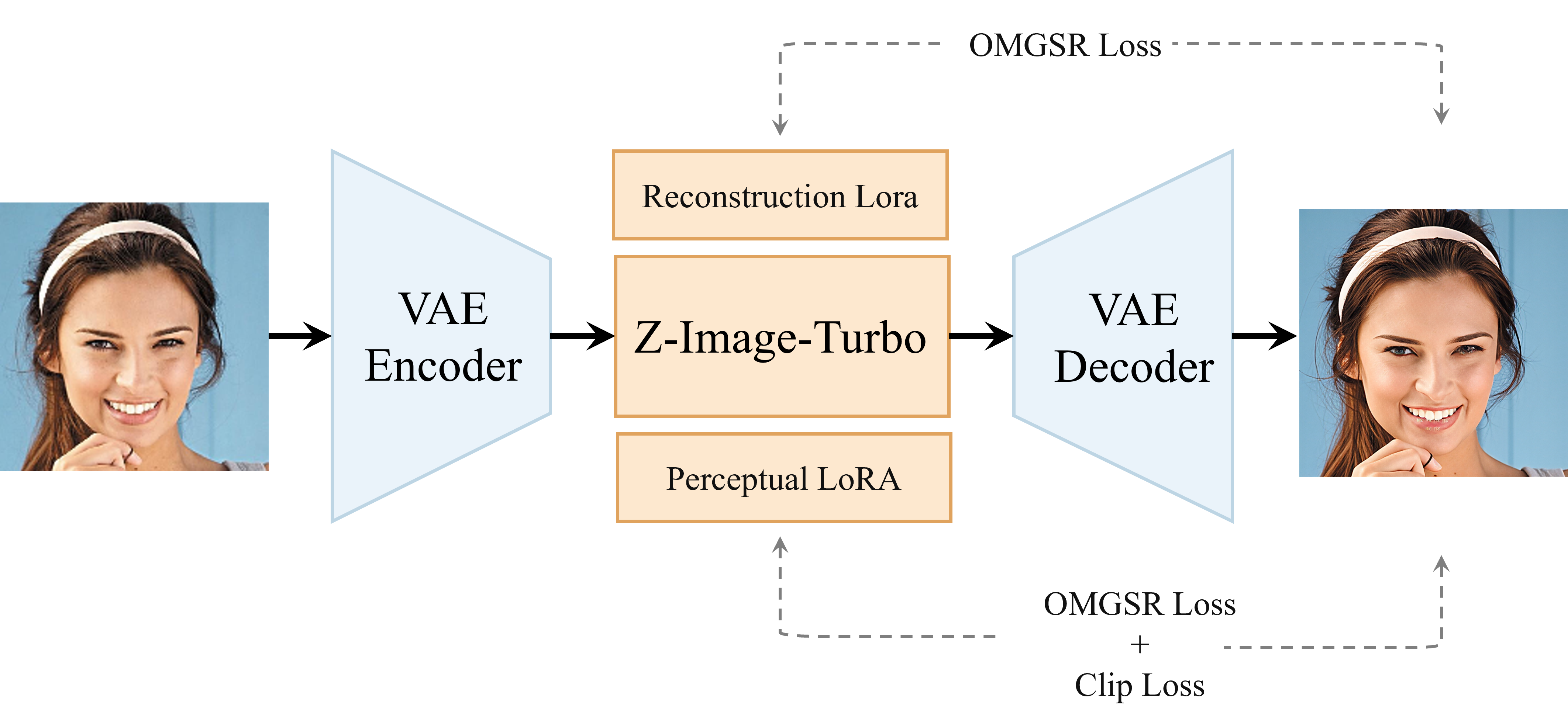}
    \vspace{-5mm}
    \caption{Overview of the HONORAICamera pipeline.}
    \label{fig:team02-pipeline}
    \vspace{-2mm}
\end{figure}

\vspace{0.5mm}
\noindent\textit{Training and optimization.} As it is illustrated in Fig.~\ref{fig:team06-pipeline}, only the SA-FGRC module is trained, while the three backbone stages remain frozen. The loss combines high-frequency reconstruction, a FID-proxy term, ArcFace identity supervision~\cite{deng2019arcface}, and LPIPS perceptual loss~\cite{zhang2018lpips}. The team trains on 399 FFHQ images with precomputed stage-2 outputs using AdamW with learning rate $1\times10^{-4}$, cosine annealing for 30 epochs, and batch size 4.

\begin{figure*}[t]
    \centering
    \includegraphics[width=0.9\textwidth]{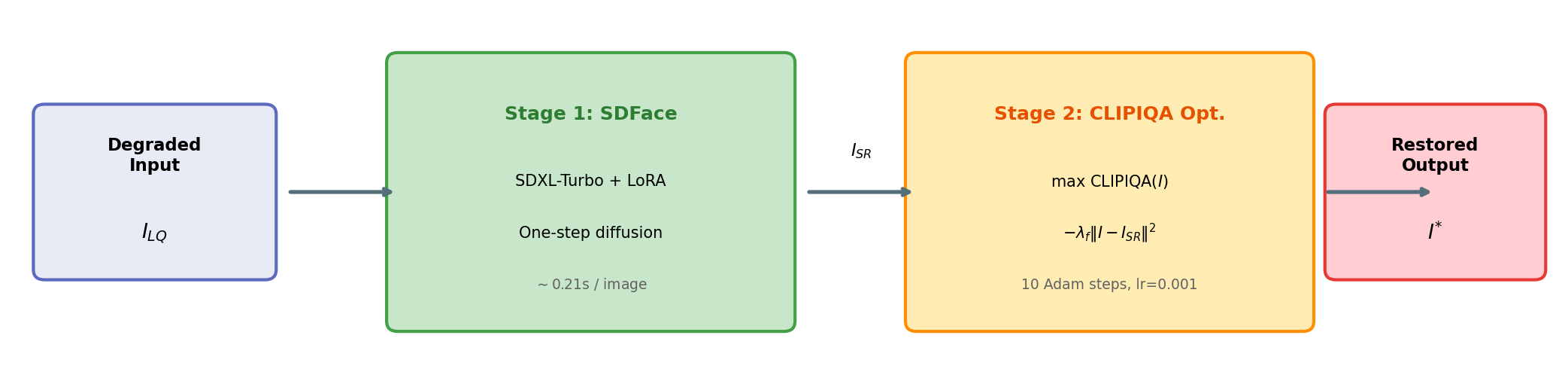}
    \vspace{-1mm}
    \caption{YuFans combines a one-step SDFace restoration stage with CLIPIQA-guided pixel optimization at test time.}
    \label{fig:team08-pipeline}
    \vspace{2mm}
\end{figure*}

\begin{figure*}[t]
    \centering
    \includegraphics[width=0.98\linewidth]{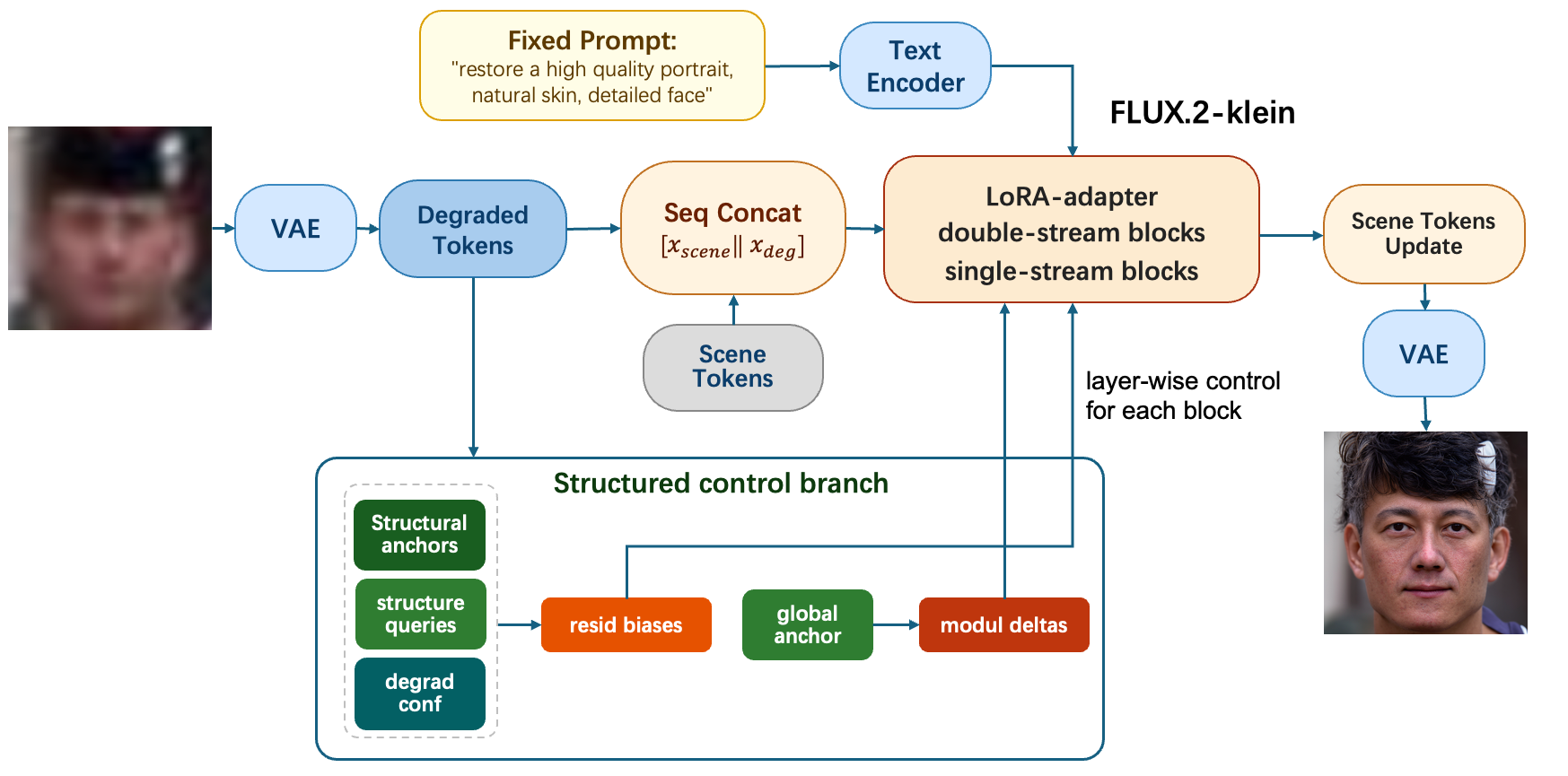}
    \vspace{2mm}
    \caption{Overview of DeSC-Face. The degraded image is encoded into degraded latent tokens, which are used both as the main condition for the LoRA-adapted FLUX.2 backbone and as the input to the structured control branch. The scene-token stream is iteratively restored and then decoded into the final output image.}
    \label{fig:team10-pipeline} \vspace{2mm}
\end{figure*}

\subsection{HONORAICamera}
\noindent\textbf{Description.} HONORAICamera builds on the diffusion-based generative prior of Z-Image-Turbo~\cite{zit} and adopts the training strategy of OMGSR~\cite{omgsr} for real-world face restoration. The method fixes both training and inference to timestep 244, aiming to balance reconstruction quality and computational efficiency in a one-step generative setting.

\vspace{1mm}
\noindent\textbf{Implementation Details.} The team synthesizes training pairs with the Real-ESRGAN degradation pipeline, including blur kernels, Gaussian and Poisson noise, and JPEG compression. The resulting training process transfers the generative prior of Z-Image-Turbo to the restoration task while keeping the output aligned with the challenge resolution and portrait content, as shown in Fig.~\ref{fig:team02-pipeline}.

\vspace{1mm}
\noindent\textit{Two-stage training.} The first stage performs generative-prior transfer at 1,024$\times$1,024 on LSDIR, FFHQ, DIV2K, and Flickr2K\_train with a total batch size 128. The second stage fine-tunes the model on FFHQ at $512\times512$, which matches the test resolution, and optimizes MSE, Dv3D, GAN, and LRR losses together with an additional CLIP loss that explicitly targets higher perceptual scores.

\subsection{YuFans}
\noindent\textbf{Description.} YuFans proposes a two-stage pipeline that combines one-step diffusion restoration with test-time IQA-guided pixel optimization. Stage 1 uses SDFace~\cite{wang2025osdface}, a one-step SDXL-Turbo-based face restorer, while Stage 2 directly optimizes the restored pixels with differentiable CLIPIQA~\cite{wang2022clipiqa} under strong fidelity regularization.

\vspace{0.5mm}
\noindent\textbf{Implementation Details.} Shown in Fig.~\ref{fig:team08-pipeline}, the first stage produces an initial restoration with a pre-trained SDFace model. The second stage treats that result as initialization and performs 10 Adam gradient-ascent steps with learning rate 0.001 to maximize CLIPIQA while penalizing deviation from the SDFace output and enforcing total-variation smoothness. The fidelity and TV weights are set to $\lambda_f=20.0$ and $\lambda_{tv}=0.001$, respectively.

\vspace{0.5mm}
\noindent\textit{Training and optimization.} YuFans does not further train the base restorer. The team directly uses the pre-trained SDFace checkpoint originally trained on FFHQ~\cite{karras2019ffhq} and performs only test-time optimization in the second stage. CLIPIQA is implemented with the PyIQA toolbox~\cite{pyiqa}.

\subsection{guaguagua}
\noindent\textbf{Description.} The guaguagua team updates its submission to \emph{DeSC-Face}, short for Degradation-Aware Structured Control for Blind Face Restoration. Built on the official FLUX.2-klein-4B checkpoint, the method encodes the degraded input into latent tokens and uses them in two ways: as the main condition for the restoration backbone and as the input to a dedicated structured control branch. A separate scene-token stream is then iteratively restored and decoded into the final face image.

\vspace{0.5mm}
\noindent\textbf{Implementation Details.} As shown in Fig.~\ref{fig:team10-pipeline}, DeSC-Face concatenates scene tokens and degraded tokens along the sequence dimension before passing them through a LoRA-adapted FLUX.2 transformer. Its degradation-aware controller performs local smoothing and residual decomposition on degraded latents, then extracts structural anchors, structure queries, and degradation confidence. These signals are injected into the backbone as token-wise residual biases and modulation offsets, enabling the restoration trajectory to adapt to the estimated corruption pattern while keeping the scene stream as the only decoded output.

\vspace{0.5mm}
\noindent\textit{Training and inference.} The updated factsheet states that the submission is trained only on FFHQ and synthetically degraded FFHQ counterparts, without external data. Optimization uses LoRA rank 16, mixed precision bf16, 10 epochs, batch size 2 per device with gradient accumulation, learning rate $5\times10^{-5}$, and degradation scale range 0 to 16. Inference processes the five official test subsets independently with a fixed restoration prompt and seed 42, and reports 23.06 seconds per image in the official wrapper.

\subsection{NTR}
\begin{figure}[t]
\centering
\resizebox{\columnwidth}{!}{%
\begin{tikzpicture}[
    node distance=0.8cm and 1.0cm,
    block/.style={rectangle, draw=black!70, fill=blue!8, minimum height=0.9cm, minimum width=2.0cm, align=center, rounded corners=2pt, font=\small},
    stage/.style={rectangle, draw=black!50, fill=orange!8, minimum height=0.9cm, minimum width=2.0cm, align=center, rounded corners=2pt, font=\small},
    io/.style={rectangle, draw=black!40, fill=gray!10, minimum height=0.9cm, minimum width=1.6cm, align=center, rounded corners=2pt, font=\small},
    arr/.style={-{Stealth[length=3mm]}, thick, black!70},
]
\node[io] (input) {Degraded\\Face $X_{\text{LQ}}$};
\node[block, right=of input] (swinir) {SwinIR\\(Stage 1)};
\node[io, right=of swinir] (clean) {$\hat{X}_{\text{clean}}$};
\node[stage, right=1.2cm of clean] (controlnet) {IRControlNet\\(Stage 2)};
\node[block, above=0.5cm of controlnet] (sd) {Stable Diffusion\\2.1};
\node[io, right=of controlnet] (output) {Restored\\Face $\hat{X}$};
\draw[arr] (input) -- (swinir);
\draw[arr] (swinir) -- (clean);
\draw[arr] (clean) -- (controlnet);
\draw[arr] (controlnet) -- (output);
\draw[arr] (sd) -- (controlnet);
\begin{scope}[on background layer]
  \node[fit=(swinir), fill=blue!5, draw=blue!30, dashed, inner sep=4pt, label={[font=\scriptsize,blue!60]below:Degradation Removal}] {};
  \node[fit=(controlnet)(sd), fill=orange!5, draw=orange!30, dashed, inner sep=4pt, label={[font=\scriptsize,orange!60]below:Texture Synthesis}] {};
\end{scope}
\end{tikzpicture}
}
\caption{Architecture diagram of the DiffBIR v2.1 two-stage pipeline used by NTR. Stage 1 removes degradations with SwinIR, and Stage 2 synthesizes facial textures with IRControlNet conditioned on Stable Diffusion 2.1.}
\label{fig:team01-pipeline}\vspace{-2mm}
\end{figure}
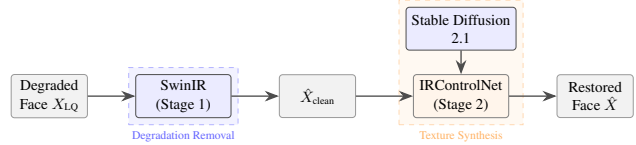

\noindent\textbf{Description.} NTR directly adopts the pre-trained DiffBIR v2.1~\cite{diffbir} model as its restoration backbone. DiffBIR is a two-stage blind image restoration framework that combines regression-based degradation removal with a generative diffusion prior for realistic texture synthesis, as illustrated by the updated architecture diagram in Fig.~\ref{fig:team01-pipeline}.

\begin{figure*}[h]
    \centering
    \includegraphics[width=\linewidth]{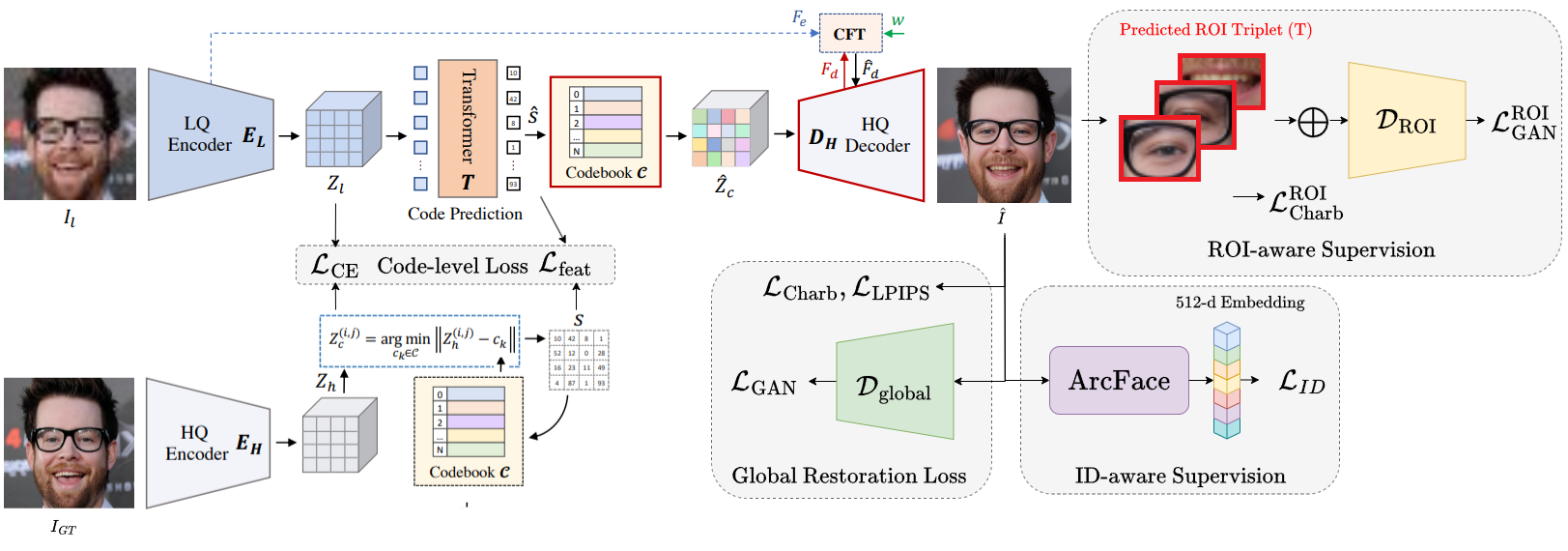}
    \vspace{-6mm}
    \caption{Overall architecture and training objective of MaDENN. The baseline CodeFormer architecture is extended with identity-preserving and ROI-aware supervision, while low-quality inputs are synthesized with a second-order Real-ESRGAN degradation pipeline.}
    \label{fig:pipeline}
        \vspace{-4mm}
\end{figure*}

\vspace{0.5mm}
\noindent\textbf{Implementation Details.} The first stage is a SwinIR~\cite{swinir}-based cleaning module that removes blur, noise, and compression artifacts and outputs a coarse clean estimate. The second stage is an IRControlNet built on Stable Diffusion 2.1~\cite{rombach2022ldm,zhang2023controlnet}, which conditions on the coarse estimate and synthesizes high-frequency facial textures.

\vspace{0.5mm}
\noindent\textit{Training and inference.} The team uses the released DiffBIR v2.1 weights without additional fine-tuning. Inference uses the EDM DPM++\,3M SDE sampler~\cite{karras2022edm} with 10 diffusion steps, guidance scale 6.0, strength 1.0, FP16 precision, and random seed 231. The factsheet reports roughly 1.0--1.5 seconds per image on a single NVIDIA H100 GPU.


\subsection{MaDENN}
\noindent\textbf{Description.} MaDENN builds upon CodeFormer~\cite{zhou2022codeformer} and focuses on strengthening the training methodology. Their solution combines second-order Real-ESRGAN degradation synthesis~\cite{realesrgan}, ArcFace-based identity preservation~\cite{deng2019arcface}, and ROI-aware supervision on semantically critical facial components, as illustrated in Fig.~\ref{fig:pipeline}.

\vspace{0.5mm}
\noindent\textbf{Implementation Details.} Compared with the original CodeFormer training recipe, MaDENN enlarges the degradation space with a second-order Real-ESRGAN process and adds two extra supervision sources. The first is an ArcFace identity loss that keeps restored faces close to the ground-truth identity embedding. The second is a triplet ROI loss on the left eye, right eye, and mouth, where RoIAlign-based crops and a dedicated ROI discriminator encourage better local structure and bilateral symmetry.

\vspace{0.5mm}
\noindent\textit{Training setup.} The team fine-tunes a public CodeFormer checkpoint on FFHQ~\cite{karras2019ffhq}, excluding samples from the FFHQ-Ref-Test split~\cite{RefLDM}. The codebook and HQ decoder remain frozen, while the LQ encoder, transformer, and controllable feature transformation layers stay trainable. Optimization uses AdamW with learning rate $1\times10^{-5}$ for 500K iterations and batch size 4.
\begin{figure}[t]
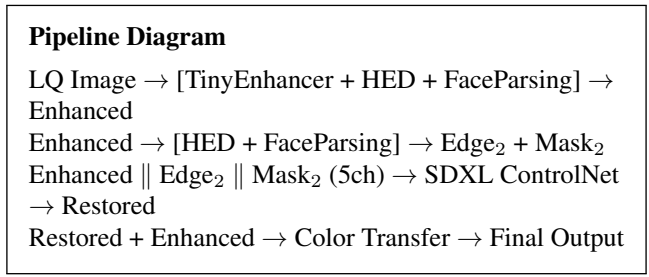

    \centering
    \setlength{\fboxsep}{8pt}
    \setlength{\fboxrule}{0.5pt}
    \fbox{%
        \begin{minipage}{0.95\columnwidth}
            \raggedright
            \textbf{Pipeline Diagram}

            \vspace{0.4em}
            LQ Image $\rightarrow$ [TinyEnhancer + HED + FaceParsing] $\rightarrow$ Enhanced\\
            Enhanced $\rightarrow$ [HED + FaceParsing] $\rightarrow$ Edge$_2$ + Mask$_2$\\
            Enhanced $\Vert$ Edge$_2$ $\Vert$ Mask$_2$ (5ch) $\rightarrow$ SDXL ControlNet $\rightarrow$ Restored\\
            Restored + Enhanced $\rightarrow$ Color Transfer $\rightarrow$ Final Output
        \end{minipage}%
    }
    \vspace{-0.8mm}
    \caption{The SN VISION pipeline first enhances the degraded face with TinyEnhancer and auxiliary structural cues, then feeds the enhanced RGB image together with refined edge and face-mask maps into SDXL ControlNet for final restoration.}
    \label{fig:team09-pipeline}\vspace{-3.3mm}
\end{figure}
\subsection{SN VISION}
\noindent\textbf{Description.} SN VISION presents \emph{SDXL ControlNet with TinyEnhancer}, a two-pass face restoration pipeline that combines lightweight face-aware preprocessing with diffusion-based generation. In the first pass, TinyEnhancer restores a cleaner intermediate image with auxiliary edge and face-mask cues. In the second pass, the enhanced RGB image together with refined edge and parsing maps forms a 5-channel ControlNet condition for SDXL-based generation~\cite{zhang2023controlnet,SDXL}.

\vspace{0.5mm}
\noindent\textbf{Implementation Details.} TinyEnhancer is a U-Net-style model with channel and spatial attention, gated fusion, an OutputRefiner module, and an adaptive Gaussian blur preprocessor. It takes a 5-channel input consisting of RGB, face mask, and edge map. The second pass re-extracts HED edges and face parsing masks from the enhanced image, concatenates them with the RGB output, and feeds the resulting 5-channel tensor into an SDXL ControlNet. The team adopts txt2img rather than img2img so that LQ artifacts are not directly propagated into the diffusion process, and finishes with Reinhard LAB color transfer.

\vspace{0.5mm}
\noindent\textit{Training and inference.} All components are trained on FFHQ~\cite{karras2019ffhq} with synthetic degradations at $1024\times1024$. The ControlNet branch is initialized from DreamshaperXL v2.1 Turbo with zero-initialized weights for the extra conditioning channels and optimized with AdamW at a learning rate of $1\times10^{-5}$, batch size 4, for 135k steps. TinyEnhancer is trained with L1, perceptual, and adversarial losses; the HED branch is fine-tuned from pretrained weights; and the face parsing branch is trained as a U-Net segmentation model. At inference time, the team uses 50 diffusion steps with seed 42 and tunes several generation parameters separately for each test subset.

\subsection{ALLCAN}
\noindent\textbf{Description.} ALLCAN proposes PRIDE-Face, a two-stage framework built on DiffBIR~\cite{diffbir}. The 1st stage replaces the default restoration module with GFPGAN~\cite{wang2021gfpgan} to extract a stronger facial structural prior. The 2nd stage uses diffusion to synthesize realistic high-frequency facial details.

\begin{figure}[t]
    \centering
    \includegraphics[width=0.9\linewidth]{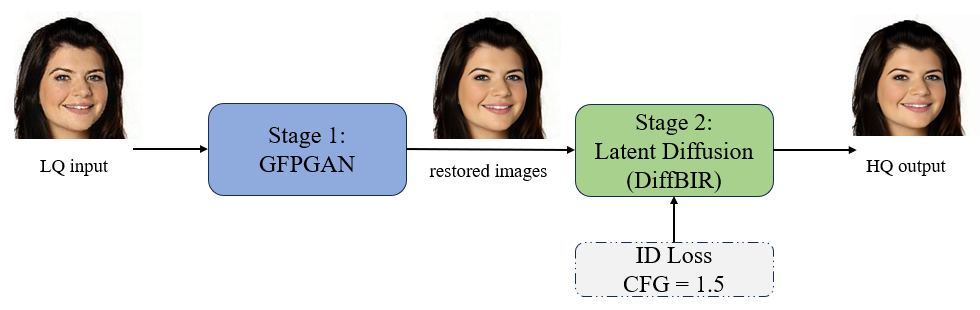}
    \vspace{-2mm}
    \caption{The workflow of PRIDE-Face. GFPGAN provides the structural prior in the first stage, while DiffBIR synthesizes high-fidelity details in the second stage.}
    \label{fig:team04-pipeline}\vspace{-2mm}
\end{figure}

\vspace{1mm}
\noindent\textbf{Implementation Details.} PRIDE-Face treats the GFPGAN output only as an intermediate spatial condition, rather than the final restored result, because the team found that direct GFPGAN outputs tend to over-smooth textures. To better preserve identity during diffusion generation, the method adds an explicit identity loss based on face-recognition embeddings between the generated result and the input image.

\vspace{1mm}
\noindent\textit{Guidance calibration.} The team further fixes the classifier-free guidance scale at 1.5 to suppress overly aggressive high-frequency hallucinations and improve perceptual naturalness. This calibrated setting is used together with the stronger first-stage structural prior to balance realism and identity consistency.

\subsection{BVI}
\noindent\textbf{Description.} BVI builds on the Time-Aware one-step Diffusion Network for real-world image super-resolution (TADSR)~\cite{tadsr}. Their main modification is a residual noise refiner inserted into the one-step student branch, together with a detail-aware training strategy that strengthens local high-frequency restoration while keeping the efficiency of one-step diffusion.

\begin{figure}[t]
    \centering
    \includegraphics[width=\columnwidth]{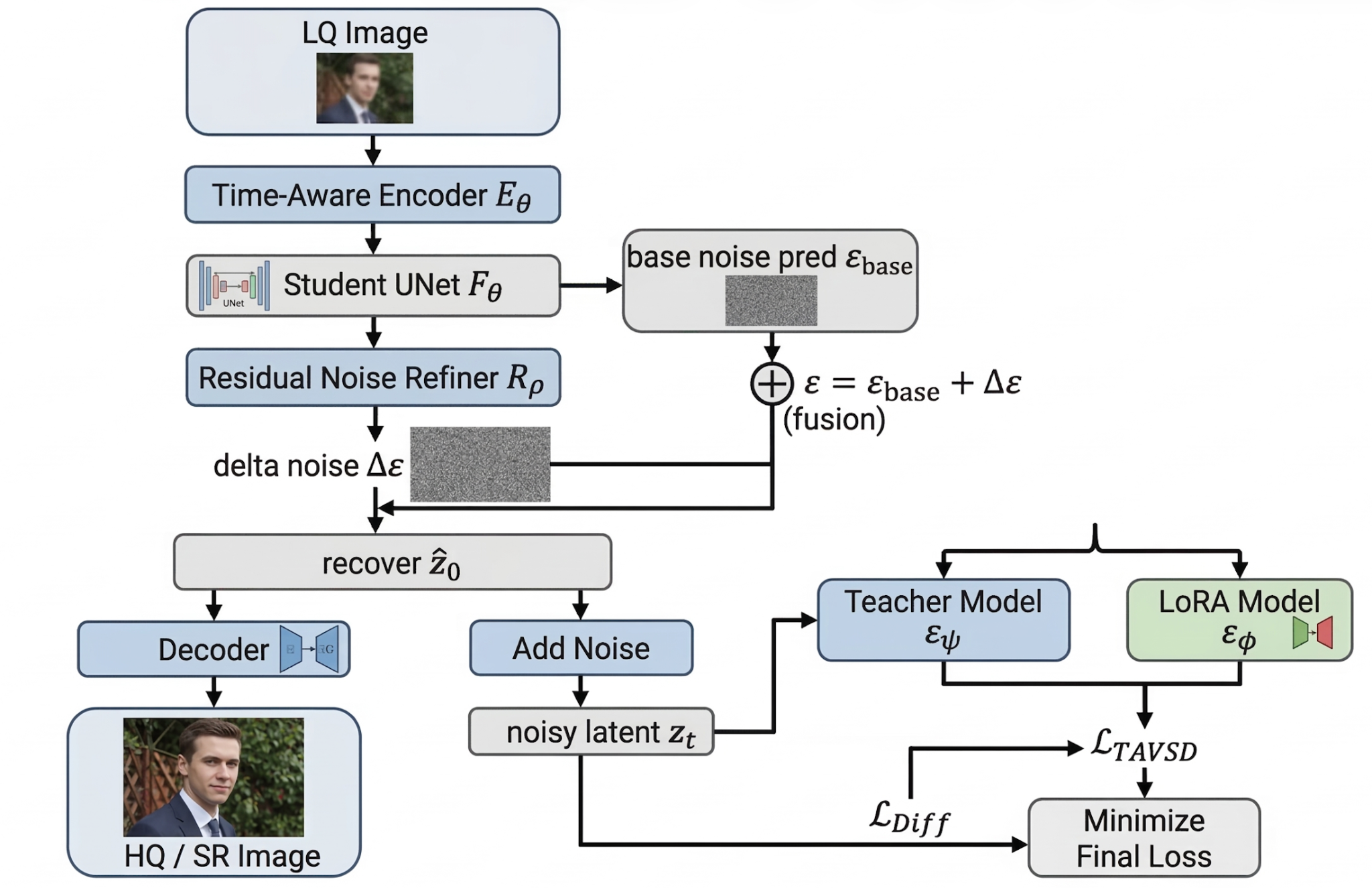}
    \caption{The BVI extends TADSR with a residual noise refiner and a detail-aware training strategy inside the one-step student branch.}
    \label{fig:team07-pipeline}
\end{figure}

\vspace{1mm}
\noindent\textbf{Implementation Details.} The time-aware encoder and student branch first predict a base noise estimate from the low-quality input and student timestep. A lightweight residual refiner then predicts a corrective term that is added to the base noise prediction before decoding. The corrected latent is forwarded to the frozen teacher and LoRA branch following the original TADSR design. To emphasize local structure, the team adds weighted Charbonnier losses on high-frequency residuals and image gradients, together with a ratio-capped regularizer that prevents the refinement branch from overwhelming the student prediction.

\vspace{1mm}
\noindent\textit{Training setup.} The method follows the TADSR training recipe with LSDIR~\cite{li2023lsdir}, BVI-AOM~\cite{nawala2024bvi}, FFHQ-style data~\cite{karras2019ffhq}, and Real-ESRGAN degradation synthesis~\cite{realesrgan}. The challenge setting uses a scale factor of 1 and retains the original diffusion-prior training formulation of TADSR.

\vspace{-2.mm}
\section*{Acknowledgements}
\vspace{-2.mm}
This work is supported by the National Natural Science Foundation of China (62501386, 625B2116, 625B1025), CCF-Tencent Rhino-Bird Open Research Fund. This work is also sponsored by Al Hundred Schools Program and is carried out using the Ascend AI technology stack.
This work is partially supported by the Humboldt Foundation. We thank the NTIRE 2026 sponsors: OPPO, Kuaishou, and the University of Wurzburg (Computer Vision Lab).

\newpage
\appendix

\section{Teams and Affiliations}
\label{sec:app_teams}

\subsection*{MiPlusCV}
\noindent\textit{\textbf{Title: }} Two-Stage OSDFace and Z-Image Face Restoration

\noindent\textit{\textbf{Members: }} \\
Wei Deng\textsuperscript{1}(\href{mailto:dengwei1@xiaomi.com}{dengwei1@xiaomi.com}), WenBo Xiong\textsuperscript{1}, Yifei Chen\textsuperscript{1}, Xian Hu\textsuperscript{1}, Daiguo Zhou\textsuperscript{1}

\noindent\textit{\textbf{Affiliations: }}\\
\textsuperscript{1}MiLM Plus, Xiaomi Inc., China

\subsection*{CEVI-KLETech}
\noindent\textit{\textbf{Title: }} Semantic-Aware Wavelet Frequency Refiner for Face Restoration

\noindent\textit{\textbf{Members: }} \\
Nikhil Akalwadi\textsuperscript{1}(\href{mailto:nikhil.akalwadi@kletech.ac.in}{nikhil.akalwadi@kletech.ac.in}), Sujith Roy V\textsuperscript{1}, Claudia Jesuraj\textsuperscript{1}, Vikas B\textsuperscript{1}, Spoorthi LC\textsuperscript{1}, Ramesh Ashok Tabib\textsuperscript{1}, Uma Mudenagudi\textsuperscript{1}

\noindent\textit{\textbf{Affiliations: }}\\
\textsuperscript{1}KLE Technological University, Hubballi, India

\subsection*{HONORAICamera}
\noindent\textit{\textbf{Title: }} Diffusion-based Generative Prior for Real-World Face Restoration

\noindent\textit{\textbf{Members: }} \\
Yingsi Chen\textsuperscript{1}(\href{mailto:chenyingsi@honor.com}{chenyingsi@honor.com}), Yijiao Liu\textsuperscript{1}, Hui Li\textsuperscript{1}, Yu Wang\textsuperscript{1}, Congchao Zhu\textsuperscript{1}

\noindent\textit{\textbf{Affiliations: }}\\
\textsuperscript{1}Honor Device Co. Ltd

\subsection*{YuFans}
\noindent\textit{\textbf{Title: }} SDFace with CLIPIQA-Guided Pixel Optimization

\noindent\textit{\textbf{Members: }} \\
Wei Zhou\textsuperscript{1}(\href{mailto:weichow@u.nus.edu}{weichow@u.nus.edu}), Linfeng Li\textsuperscript{1}, Hongyu Huang\textsuperscript{2}

\noindent\textit{\textbf{Affiliations: }}\\
\textsuperscript{1}National University of Singapore\\
\textsuperscript{2}Zhejiang University

\subsection*{guaguagua}
\noindent\textit{\textbf{Title: }} DeSC-Face: Degradation-Aware Structured Control for Blind Face Restoration

\noindent\textit{\textbf{Members: }} \\
Axi Niu\textsuperscript{1}, Jinyang Zhang\textsuperscript{1}(\href{mailto:zhangjinyang@mail.nwpu.edu.cn}{zhangjinyang@mail.nwpu.edu.cn}), Zhenguo Wu\textsuperscript{1}, Senyan Qing\textsuperscript{1}

\noindent\textit{\textbf{Affiliations: }}\\
\textsuperscript{1}Northwestern Polytechnical University, China

\subsection*{NTR}
\noindent\textit{\textbf{Title: }} DiffBIR v2.1 for Real-World Face Restoration

\noindent\textit{\textbf{Members: }} \\
Jiachen Tu\textsuperscript{1}(\href{mailto:jtu9@illinois.edu}{jtu9@illinois.edu}), Guoyi Xu\textsuperscript{1}, Yaoxin Jiang\textsuperscript{1}, Jiajia Liu\textsuperscript{1}, Yaokun Shi\textsuperscript{1}

\noindent\textit{\textbf{Affiliations: }}\\
\textsuperscript{1}University of Illinois Urbana-Champaign

\subsection*{MaDENN}
\noindent\textit{\textbf{Title: }} Identity-Preserving CodeFormer with ROI-Aware Supervision

\noindent\textit{\textbf{Members: }} \\
Alexandru-Gabriel Lefterache\textsuperscript{1}(\href{mailto:alefterache@upb.ro}{alefterache@upb.ro}), Anamaria Radoi\textsuperscript{1}

\noindent\textit{\textbf{Affiliations: }}\\
\textsuperscript{1}UNSTPB POLITEHNICA Bucharest, Romania

\subsection*{SN VISION}
\noindent\textit{\textbf{Title: }} SDXL ControlNet with TinyEnhancer: A Two-Pass Pipeline for Face Restoration

\noindent\textit{\textbf{Members: }} \\
Hoyoung Lee\textsuperscript{1}(\href{mailto:hoyounglee@snowcorp.com}{hoyounglee@snowcorp.com}), SangYun Oh\textsuperscript{1}, ChangYoung Jeong\textsuperscript{1}

\noindent\textit{\textbf{Affiliations: }}\\
\textsuperscript{1}SNOW Corporation

\subsection*{ALLCAN}
\noindent\textit{\textbf{Title: }} PRIDE-Face

\noindent\textit{\textbf{Members: }} \\
Chuanyue Yan\textsuperscript{1}(\href{mailto:chuanyueyan0820@163.com}{chuanyueyan0820@163.com}), Tao Lu\textsuperscript{1}, Yanduo Zhang\textsuperscript{1}, Kanghui Zhao\textsuperscript{1}, Jiaming Wang\textsuperscript{1}, Yuqi Li\textsuperscript{2}

\noindent\textit{\textbf{Affiliations: }}\\
\textsuperscript{1}Wuhan Institute of Technology\\
\textsuperscript{2}City University of New York

\subsection*{BVI}
\noindent\textit{\textbf{Title: }} TADSR with Residual Noise Refinement for Face Restoration

\noindent\textit{\textbf{Members: }} \\
Yuxuan Jiang\textsuperscript{1}(\href{mailto:dd22654@bristol.ac.uk}{dd22654@bristol.ac.uk}), Chengxi Zeng\textsuperscript{1}, Tianhao Peng\textsuperscript{1}, Fan Zhang\textsuperscript{1}, David Bull\textsuperscript{1}

\noindent\textit{\textbf{Affiliations: }}\\
\textsuperscript{1}University of Bristol

{\small
\bibliographystyle{ieeenat_fullname}
\bibliography{main}
}

\end{document}